\pdfoutput=1

\documentclass[11pt]{article}

\usepackage[]{acl}
\usepackage{times}
\usepackage{latexsym}

\usepackage[T1]{fontenc}

\usepackage[utf8]{inputenc}

\usepackage{graphicx}
\usepackage{microtype}
\usepackage{amsmath}
\usepackage{xcolor}
\usepackage{comment}
\usepackage{soul}

\usepackage{soul}
\newcommand\cl[1]{\textcolor{green}{#1}}

\usepackage{booktabs}
\usepackage{enumitem}
\newcommand{\ra}[1]{\renewcommand{\arraystretch}{#1}}

\newenvironment{myitemize}{\begin{itemize}[noitemsep,topsep=0pt,leftmargin=*,label={}]}{\end{itemize}}
\title{\textsc{milIE}: Modular \& Iterative Multilingual Open Information Extraction}

 \author{Bhushan Kotnis$^{1}$, Kiril Gashteovski$^{1}$, Daniel Oñoro-Rubio$^{1}$,\\
 {\bf Ammar Shaker$^{1}$, Vanesa Rodriguez-Tembras$^{2}$, Makoto Takamoto$^{1}$,}\\ {\bf Mathias Niepert$^{1,3}$, Carolin Lawrence$^{1}$}  \\
         $^{1}$NEC Laboratories Europe, Heidelberg, Germany.\\ \texttt{firstname.lastname@neclab.eu} \\ $^{2}$Heidelberg University, Center for Ibero-American Studies, Germany. \\
         $^{3}$University of Stuttgart, Germany}

\begin{document}
\maketitle
\begin{abstract}
Open Information Extraction (OpenIE) is the task of extracting \emph{(subject, predicate, object)} triples from natural language sentences. Current OpenIE systems extract all triple slots independently. In contrast, we explore the hypothesis that it may be beneficial to extract triple slots iteratively: first extract easy slots, followed by the difficult ones by conditioning on the easy slots, and therefore achieve a better overall extraction.

Based on this hypothesis, we propose a neural OpenIE system, \textsc{milIE}, that operates in an iterative fashion. 
Due to the iterative nature, the system is also modular\textemdash it is possible to seamlessly integrate rule based extraction systems with a neural end-to-end system, thereby allowing rule based systems to supply extraction slots which \textsc{milIE} can leverage for extracting the remaining slots.
We confirm our hypothesis empirically: \textsc{milIE} outperforms SOTA systems on multiple languages ranging from Chinese to Arabic. Additionally, we are the first to provide an OpenIE test dataset for Arabic and Galician. 

\end{abstract}

\section{Introduction}

Open Information Extraction (OpenIE) aims to extract structured facts in the form of \emph{(subject, relation, object)}-triples from natural language sentences \cite{Etzioni2008}.  For example, given a sentence, \textit{"Barrack Obama became the US President in the year 2008"}, an OpenIE system is expected to extract the following triples: (\textit{Barrack Obama}; \textit{became}; \textit{US President}) and (\textit{Barrack Obama}; \textit{became US President in}; \textit{2008}). We refer to subject, predicate and the object of the triple as slots of a triple. 
OpenIE extractions are schema-free, human understandable intermediate representations of facts in source texts \cite{Mausam2016}. They are useful in a variety of information extraction end tasks such as summarization \cite{xu2021generating}, question answering \cite{Khot2017,yan2018assertion} and automated schema extraction \cite{Nimishakavi2016}.

The various slots of a triple are dependent on each other and hence an error in one slot renders the entire extraction unusable. We hypothesize that triple extraction errors largely stem from the difficulty of extracting certain slots of a triple and said difficulty  may depend on the sentence construction and the language. For example, \textit{"Barrack Obama became the US President in the year 2008"} contains two triples \textit{(Barrack Obama; became; US President)} and \textit{(Barrack Obama; became US President in; 2008)}. Extracting the predicate, \textit{"became US President in"}, for the second triple is tricky, because the object of the first triple (US President) overlaps with the predicate of the second triple. But if the extraction system was provided with the object, \textit{(2008)}, and then asked to extract a triple conditioned on this object, the predicate extraction would be easier.

This is precisely the hypothesis we wish to investigate \textemdash is it easier to extract certain slots of a triple, say subjects, compared to other slots, such as objects, and is it possible to improve performance by leveraging specific slot extraction orders?


Given the hypothesis, we propose \textsc{milIE}, a \textsc{m}odular \& \textsc{i}terative multi\textsc{l}ingual open \textsc{I}nformation \textsc{E}xtraction system, 
which iteratively extracts the different slots of a triple. The iterative nature allows for (1) studying the effects of a slot extractions on the remaining extractions, (2) extracting easier triple slots followed by harder ones, (3) aggregating different slot extraction orders as a mixture of experts, and (4) integrating slots supplied by an external rule-based system, resulting in a hybrid system. The latter offers a system that combines the best of neural and rule based systems, e.g. by using a rule-based system to extract high precision slots on which the neural system is conditioned.

We empirically confirm our hypothesis: the iterative nature of \textsc{milIE} outperforms several SOTA systems. It proves especially useful for zero-shot multilingual extraction, which we evaluated on five different low resource languages.
Additionally we show how \textsc{milIE} can leverage rule-based slot extraction by
conditioning on them to predict the remaining parts of the triple. Therefore \textsc{milIE} is a boon for existing applications wishing to transition from a rule based information extraction system to a neural one, because \textsc{milIE} would allow using the rule-based system to compensate for the lack of exhaustive training data. Finally, we perform linguistic analyses that uncovers useful insights on how different languages either make it easy or difficult for OpenIE systems to extract individual elements of the triple.

Our contributions are summarized as follows:
\begin{enumerate}
	\item We propose \textsc{milIE}, a multilingual OpenIE system that iteratively extracts the different slots of a triple. 
	\item We carry out extensive experiments on a variety of languages (English, Chinese, German, Arabic, Galician, Spanish and Portuguese) and demonstrate that \textsc{milIE} outperforms recent SOTA systems by a wide margin, especially on languages other than English. 
	\item We perform an extensive analysis based on ablation studies and uncover interesting insights about the nature of OpenIE task in different languages.
\end{enumerate}

\section{\textsc{milIE}}


The backbone of our system is the iterative procedure (Section \ref{sec:iter}), which allows us to investigate our hypothesis. The iterative procedure allows us to extract triple slots in various pathway orders, which results in a series of possible aggregation schemes (Section \ref{sec:pathways}). To create a strong iterative system, the training paradigm (Section \ref{sec:training} needs to consider two aspects: (1) it needs to prepare incomplete triple extractions which represent incomplete triple extractions the system is expected to predict; (2) it creates negative samples that allow for teaching the system when to not continue with an extraction due to a prior error.  With the iterative nature we also integrate rule-based systems (Section \ref{sec:rule}) as well as elegantly handle the specific case of n-ary extractions, where more than 3 slots need to be extracted (Section \ref{sec:binarization}).

\subsection{Iterative Prediction}\label{sec:iter}
To implement the iterative nature of our system, we use a BERT-based transformer \cite{Devlin2019bert} as the base building block. On top of this block, we add a total of four neural networks blocks in parallel, which we refer to as heads and which are each in charge of extracting a particular triple slot. Concretely, we have the heads $f_{s}, \ f_{o}, \ f_{p}, \ f_{a}$, which are in charge of predicting \textit{s}ubject, \textit{o}bject, \textit{p}redicate and \textit{a}rgument, respectively. The argument head is an extra feature, which is needed for n-ary extractions that occur in some datasets, where in addition to the triple there might be an argument that modifies the triple, e.g., a temporal phrase.

\begin{figure}
	\includegraphics[scale=0.088]{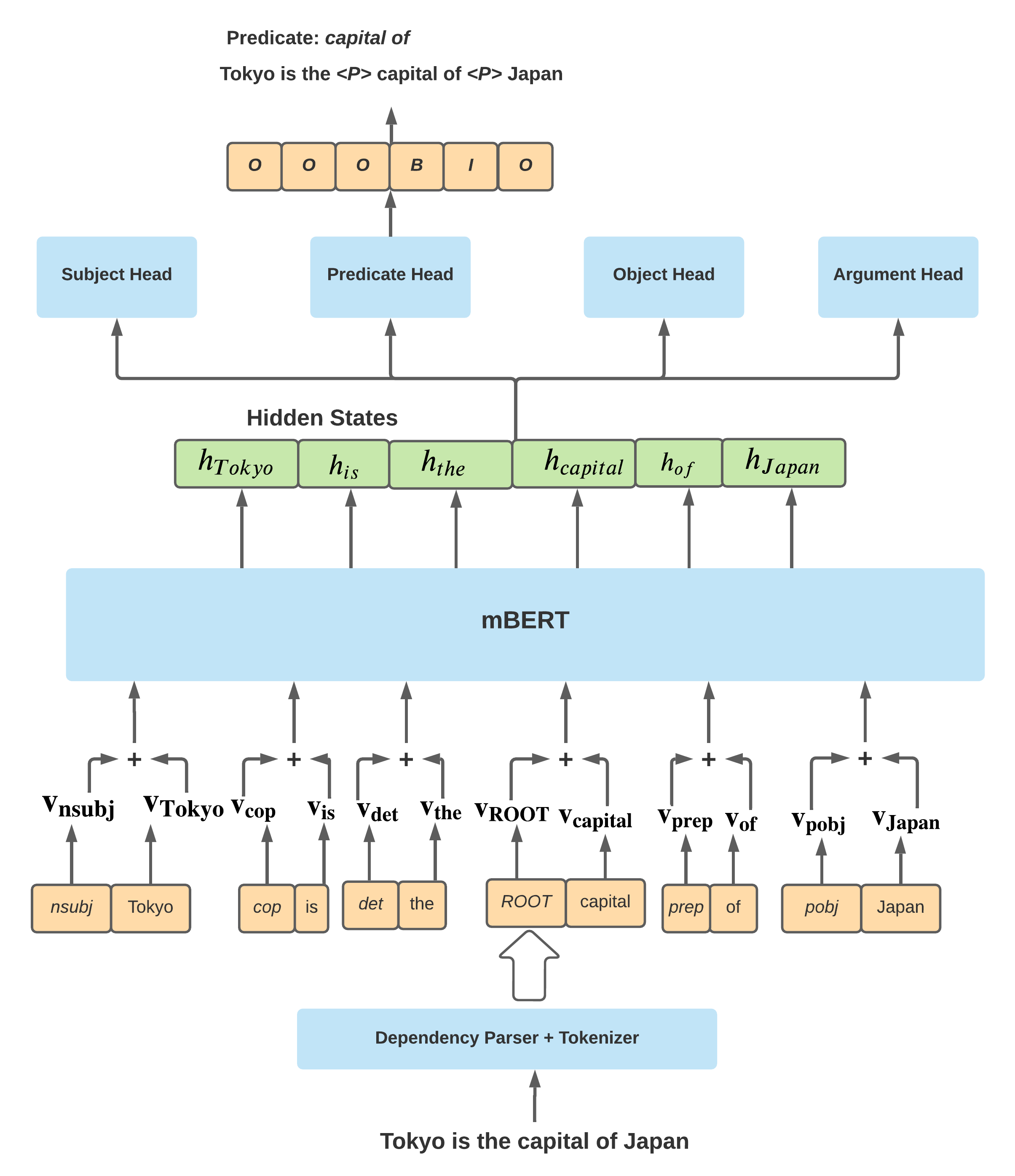}
	\caption{\textsc{milIE} system architecture. An input sequence is is tokenized and, optionally, dependency parsed. This is given to a BERT-based transformer, which outputs a hidden state for each token. The hidden states are given to each of the extraction heads, here to the predicate head. This head marks the location of the predicate in the sequence. The system then proceeds to extract the other slots, see Figure \ref{fig:extraction}.}
	\label{fig:system}
\end{figure}

Given an input sequence of words of length $N$, $S = w_1, \cdots, w_N$, the task for each extraction head is framed as a BIO tagging problem. For this, each output head outputs a label $l_i$ for token $w_i$, where $l_i \ \in \{B, I, O\}, \ i = 1 \cdots N$ ( see Figure \ref{fig:system} for the architecture). The output heads use the final transformer hidden state and predict labels denoted by $L_{s}, L_{o}, L_{p}, L_{a}$ where $L_{(\cdot)} = l_1,l_2,\ \cdots l_N$.

By having different extraction heads, we identify extraction slots iteratively. During prediction time, along with the input sentence, the model also expects extractions predicted by the previous iterations. To provide this information we add special symbols to the sentence that explicitly mark the previous extractions in the sentence. For example, we surround the predicate with the symbol \textit{<P>}, subject with \textit{<S>} and object with \textit{<O>}. For example, for predicting the object given the predicate extracted from previous iteration, the extracted predicate is marked in the sentence using the \textit{<P>} symbol and the sentence is consequently passed through the transformer for predicting the object using the object head. We always extract the arguments at the last iteration, therefore we do not mark the arguments in the sentence.\cl{\footnote{Preliminary experiments suggested that predicted the argument last leads to better overall results. This makes sense intuitively, as the argument can modify the entire triple.}}


Finally, we add the option to attach a dependency tag $t_i$ to each word $w_i$ in the sequence. This additional information may allow the system to more effectively learn how to extract triples. We use a language specific dependency tagger for obtaining the tags. We target languages, which are low resource for OpenIE, but could be high resource for other tasks, such as PoS tagging or dependency parsing. For a graphical overview of the \textsc{milIE} architecture, see Figure \ref{fig:system}.

\begin{figure}
	\centering
	\includegraphics[scale=0.09]{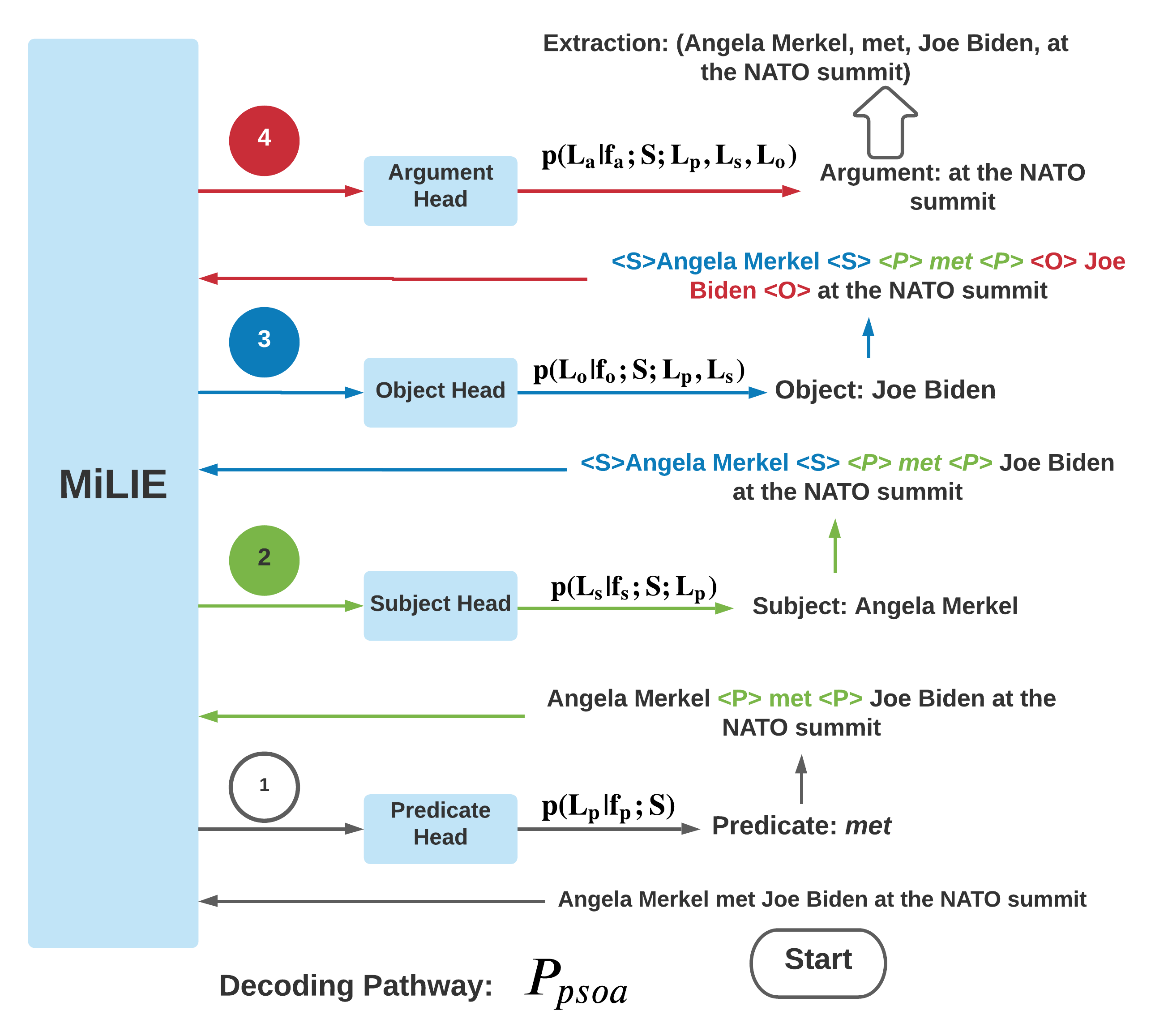}
	\caption{Iterative extraction dynamics for decoding pathway $P_{psoa}$. The numbers indicate the iteration number. Iterations are color coded, black is the predicate extraction, green subject extraction, blue object extraction and red argument extraction.} 
	\label{fig:extraction}
\end{figure}

\subsection{Aggregating Decoding Pathways}\label{sec:pathways}
The order in which the different triple parts are extracted can be varied. This allows us to investigate the challenge of extracting triple elements in specific order on different languages. Additionally different pathways aid different kinds of extractions and combining them results in a richer set of extractions. Choosing a particular order defines a decoding pathway $P_{uvxy}$ as a sequence of output heads where $u,v,x,y \in \{s,p,o,a\}$. 
For example, the decoding pathway $P_{spoa}$ denotes a sequence of output functions $(f_{s},f_{p}, f_{o}, f_{a})$. 

Fixing the n-ary argument extraction in the final iteration we obtain the following six decoding pathways- $P_{spoa}, P_{sopa}, P_{psoa},  P_{posa}, P_{ospa}, P_{opsa} $. Let's assume the decoding pathway $P_{psoa}$: predicates are extracted first, then for each predicate, subjects are extracted, then for each (predicate, subject) pair objects are extracted and finally for every extracted (predicate, subject, object) tuple all the n-ary arguments are extracted. This extraction procedure preserves the relationships between the extracted elements resulting in correctly extracting multiple triples. Figure \ref{fig:extraction} illustrates this procedure. 

We hypothesize that some triples are easier to predict if, e.g., the predicate is extracted first while for others subject first would work well. This could differ from triple to triple, but also with different languages. Consequently, some decoding pathways might be more error prone than others. This leads to two questions: (1) Which pathways are best? (2) Can we improve recall by aggregating triples using different decoding pathways?


We propose a simple algorithm we term as \textit{Water Filling} (WF) for aggregating the extractions. This is inspired by the power allocation problem in the communication engineering literature \cite{Kumar2008wireless}. Imagine a thirsty person with access to different pots of water with varying levels of purity and with the caveat that the amount of water is inversely proportional to the purity. The natural solution is to first drink the high purity water and move on to the pots in decreasing level of purity until the thirst is quenched. We use the same idea. Treating each decoding pathways as an expert, we assume that the triples extracted by all 6 pathways are more accurate compared to those extracted by only 5 pathways, 4 pathways and so on. This can be thought of as triples obtaining votes from experts. 
Starting with an empty set, for each sentence we start adding triples to the set in the order of decreasing number of received votes. The normalized votes a triple receives is used as the confidence value of the triple.  Although the procedure is explained in a sequential manner it can be parallelized by running all 6 pathways in parallel.


\subsection{Training \label{sec:training}}
\paragraph{Triple preparation.}  For effectively extracting different triple slots conditioned on other slots, the model needs to see such combinations during training. However, enumerating all possible combinations exhaustively is prohibitively expensive. We propose a sampling technique that ensures that the model sees varied combinations of different targets and prior extractions. This is done by creating a training set that simulates a prior extraction and forces the model to predict the next extraction. To ensure that the training dataset size does not explode, we randomly sample one pathway order for each training instance.

Based on the sampled pathway, we randomly sample at which step in the decoding process we are at and then mark the slots prior to this step in the sentence and use the remaining steps as target labels. We allow for multiple instances of the target labels, however there is only one instance of the marked element. For example, given one subject the target could be multiple predicates. This procedure trains the model to predict an appropriate label conditioned on a variety of previous predictions. At each time step we update the parameters of the currently used head and the underlying model.

Given that triples are at different steps in their decoding process,  we minimize different log-likelihood functions. We describe the log likelihood functions along with a few example of  the training instances in Table \ref{table:loss_functions}. We list additional details in Appendix \ref{sec:appendix}. 





\paragraph{Negative Sampling.} Iterative prediction is prone to error amplification, i.e. if an error is made during the first iteration then the error propagates and affects subsequent extractions. Anticipating this, we train \textsc{milIE} to recognize extraction errors made in the previous iteration. We purposely augment the training data with corrupted data points containing incorrectly marked extractions. For each of the incorrect extractions the model is trained to predict a blank extraction, i.e., predicting the outside label for all tokens. We use a similar sampling procedure as described previously. For every training data point from a fixed number of training data points, we create one negative sample using one of the three techniques and then choose $k$ negative samples, where $k$ is a hyperparameter. 

We corrupt triples using three techniques: (1) corrupting the predicates by replacing them with randomly chosen tokens from the sentence, (2) corrupting the subject and object by exchanging them, and (3) by mismatching the subject object pairs from different triples. We detail the entire procedure in Appendix \ref{sec:appendix}.

\begin{table*}[]
	\ra{1.1}
	\scriptsize{
		\begin{tabular}{@{}llll@{}}
			\toprule 
			\textbf{Likelihood function} & \textbf{Input Sentence} & \textbf{Head} &\textbf{Target}  \\
			\midrule
			$\mathcal{L}_p = - \sum_{i=1}^{N} log \ p( l^{p}_i  \vert  f_{p}(\theta) ; S )$ &  The Taj Mahal was built by Shah Jahan in 1643 & Predicate&built by \\
			$\mathcal{L}_s = - \sum_{i=1}^{N} log \ p( l^{s}_i  \vert  f_{s}(\theta) ;  S;  L^{p} )$ & The Taj Mahal was <P>built by<P> Shah Jahan in 1643 & Subject & Taj Mahal \\
			$\mathcal{L}_o = - \sum_{i=1}^{N} log \ p( l^{o}_i  \vert \ f_{o}(\theta) ;  S;  L^{p}; L^{s} )$ & The <S>Taj Mahal<S> was <P>built by<P> Shah Jahan in 1643 & Object & Shah Jahan \\
			$\mathcal{L}_a = - \sum_{i=1}^{N} log \ p( l^{a}_i  \vert \ f_{a}(\theta) ;  S;  L^{p}; L^{s};L^{o} )$ & The <S>Taj Mahal<S> was <P>built by<P> <O>Shah Jahan<O> in 1643 &Argument& in 1643 \\
			$\mathcal{L}_p = - \sum_{i=1}^{N} log \ p( l^{p}_i  \vert \ f_{p}(\theta) ;  S;  L^{s}; L^{o} )$ & The <S>Taj Mahal<S> was built by <O>Shah Jahan<O> in 1643. & Predicate&built by \\
			$\mathcal{L}_s = - \sum_{i=1}^{N} log \ p( l^{s}_i  \vert \ f_{s}(\theta) ;  S;  L^{o} )$ & The Taj Mahal was built by <O>Shah Jahan<O> in 1643. &Subject& Taj Mahal \\
			$\mathcal{L}_o = - \sum_{i=1}^{N} log \ p( l^{o}_i  \vert \ f_{o}(\theta) ;  S)$ & The Taj Mahal was built by Shah Jahan in 1643. &Object& Shah Jahan \\
			\bottomrule
		\end{tabular}
	}
	\caption{A few examples of training inputs and corresponding log likelihood functions. }
	\label{table:loss_functions}
\end{table*}

\begin{table*}[]
	\centering
	\ra{1.1}
	\scriptsize{
		\begin{tabular}{  p{5cm} | p{5cm} | p{5cm} }
			\toprule
			\textbf{English} & \textbf{  \cite{Ro2020multi2oie} Translation} & \textbf{Error Explanation} \\ \midrule
			The stock pot should be chilled and the solid lump of dripping which settles when chilled should be scraped clean and re-chilled for future use. & La olla de caldo debe ser enfriad\textcolor{red}{o} y la masa sólida de goteo que se asienta cuando \textcolor{blue}{[se]} enfria\textcolor{red}{da} se debe \st{raspar limpio} y re-enfriada para uso futuro. &  
			\vspace{-0.1in}
			\begin{myitemize}
				\item \textit{"enfriad\textcolor{red}{o}"}: the gender of the adjective doesn't match the noun.
				\item  "\textcolor{blue}{[se]}": missing reflexive particle.
				\item  \textit{"enfriad\textcolor{red}{a}"}: wrong use of the participle.
				\item  "\textit{\st{raspar limpio}}": syntactic error.
			\end{myitemize}
			\vspace{-0.1in}
			\\ \hline
			However, StatesWest isn't abandoning its pursuit of the much-larger Mesa. & Sin embargo, StatesWest no abandona su búsqueda de la \st{tan - Mesa grande}. & <\st{tan - Mesa grande}>: syntactically and semantically incorrect. \\ \hline
			The rest of the group reach a small shop, where Brady attempts to phone the Sheriff, but the crocodile breaks through a wall and devours Annabelle. & El resto del grupo llega a una pequeña tienda, donde Brady intent\textcolor{red}{os} \st{de tel\'efono del Sheriff}, pero \st{los saltos de cocodrilo a trav\'es de una pared}, y devora a Annabelle. &
			\vspace{-0.1in}
			\begin{myitemize}
				\item \textit{"intent\textcolor{red}{os}"}: number and the gender don't match with the noun.
				\item  \textit{"\st{de tel\'efono del Sheriff}"}: telef\'ono cannot be used as a verb.
				\item  \textit{"\st{los saltos de cocodrilo a trav\'es de una pared}"}: semantically incorrect.
			\end{myitemize}
			\vspace{-0.1in}
			\\ \hline
		\end{tabular}
		\caption{Examples of incorrectly translated sentences. Using \textcolor{red}{red} we highlight mistranslated words, using \textcolor{blue}{blue}, missing words, and with a \st{strikethrough} the parts that are semantically or syntactically incorrect.}
		\label{tab:errors}
	}
\end{table*}

\subsection{Integrating Linguistic Rule based systems}\label{sec:rule}
Crucially, each output head is conditioned on the input and the output labels extracted by the previous function. This feature allows \textsc{milIE} to seamlessly integrate rule based systems with neural systems since the conditioning can be also done on extractions obtained from rule based systems. 
This is advantageous in situations where  
a linguistic rule based system works well, for say, extracting objects. Then \textsc{milIE} can complete the missing parts of the triple conditioned on the objects. 


We treat the output of the rule based system as potential objects paired with subjects and extract the predicate connecting them. If the rule based extraction is incorrect, then \textsc{milIE} can detect the error and extract nothing. This results in more accurate extractions compared to simply post-processing the extracted tokens using linguistic rules.

\subsection{Binarizing n-ary Extractions \label{sec:binarization}}

We evaluate \textsc{milIE} on both n-ary as well as binary triple extraction datasets. One simple way to convert the n-ary extractions to binary extraction is to ignore the n-ary arguments. However, this will lead to a decrease in recall because the n-ary arguments may not be part of other extracted triples due to the initial n-ary extraction. Another method is to treat the extracted n-ary arguments as objects to the same subject, predicate pair. This would ensure that the extracted arguments are not dropped, however this may result in drop of precision since the n-ary argument may not attach to the same predicate. For example, consider the extraction (\textit{Barrack Obama; became; US President; in the year 2008}). Treating n-ary arguments as objects results in (\textit{Barrack Obama; became; US President}) and (\textit{Barrack Obama; became; in the year 2008}) resulting in an incorrect extraction.

In contrast to the above subpar solutions, the iterative nature of \textsc{milIE} allows us to elegantly address the problem of converting n-ary extractions into a binary format: we treat the extracted n-ary arguments as hypothesized objects. We then provide the extracted subject, hypothesized object pair to the model, which then extracts a new predicate conditioned on the previously extracted subject and the hypothesized object, i.e., 
$p(L_p \mid f_p(\theta); S; L_s=\textrm{"Barrack Obama"}; L_o=\textrm{"year 2008"} ).$
This creates a possibility of extracting the correct predicate, something that is not possible with existing n-ary OpenIE systems.

\section{Experiments}

\begin{table*}[]
	\centering
	\ra{1.1}
	\small{	
		\begin{tabular}{@{}lllllllllllll@{}}
			\toprule
			& \multicolumn{3}{c}{\textbf{Chinese}}         & \multicolumn{3}{c}{\textbf{German}}          & \multicolumn{3}{c}{\textbf{Arabic}}  & \multicolumn{3}{c}{\textbf{Galician}}     \\ 
			\cmidrule(lr){2-4} \cmidrule(lr){5-7} \cmidrule(lr){8-10}  \cmidrule(lr){11-13}

			& \textbf{F1} & \textbf{P} & \textbf{R} & \textbf{F1} & \textbf{P} & \textbf{R} & \textbf{F1} & \textbf{P} & \textbf{R} &\textbf{F1} & \textbf{P} & \textbf{R}    \\
			\cmidrule(lr){2-4} \cmidrule(lr){5-7} \cmidrule(lr){8-10} \cmidrule(lr){11-13} 
			\textbf{M2OIE}       &      17.1       &    25.7            & 12.8              &        4.0     &      8.9          &     2.6          &      4.9       &      16.3          & 2.9               & 8.7 & 14.7 & 6.2     \\
			
			\textbf{milIE}       &     \textbf{20.5}      &    \textbf{25.2}          & 17.3              &     8.5        &      13.4          &      6.3         &    \textemdash         &  \textemdash  &  \textemdash    & \textbf{18.3}& \textbf{23.7} & \textbf{14.8}                       \\
			\textbf{- DEP} &     19.2        &     19.8           &      \textbf{18.7}        &     8.4        &      11.3          &        6.7       &    7.3         &  \textbf{14.2}             & 4.9       & 13.9 & 16.6 & 11.9        \\ 
			\textbf{- NS}  &    17.3         &         19.6       &       15.5        &     \textbf{10.3}       & \textbf{14.3}             &  \textbf{8.0}           &     4.0        & 10.8               &  2.5           & 13.7 & 18.5 & 10.9             \\
			\textbf{- Bin} &     20.0        &      22.0          &      18.4         &       9.0      &       13.5         &     6.7  &     \textbf{7.5}        &       13.8         &  \textbf{5.1}     &       17.3     &  21.7 &  14.4 \\
			\bottomrule
		\end{tabular}
		\caption{\textsc{MiLLIE} performance comparison on multilingual BenchIE. - DEP represents \textsc{milIE} trained and evaluated without dependency tags, -NS represents absence of negative sampling, -Bin represents lack of binarizing mechanism. \textsc{milIE} always outperforms M2OIE. For Arabic no dependency tags were available, therefore the first entry for Arabic is in the line - DEP.}
		\label{tab:multilingual}
	}
\end{table*}

\begin{table*}[]
	\centering
	\ra{1.1}
	\small{	
		\begin{tabular}{@{}llllllllll@{}}
			\toprule
			&  \multicolumn{3}{c}{\textbf{Spanish (LM)}} &  \multicolumn{3}{c}{\textbf{Portuguese (LM)}}
			& \multicolumn{3}{c}{\textbf{Spanish-Clean (LM)}}       \\ 
			\cmidrule(lr){2-4} \cmidrule(lr){5-7} \cmidrule(lr){8-10}  
			
			&\textbf{F1} & \textbf{P} & \textbf{R} &\textbf{F1} & \textbf{P} & \textbf{R}  &\textbf{F1} & \textbf{P} & \textbf{R}    \\
			\cmidrule(lr){2-4} \cmidrule(lr){5-7} \cmidrule(lr){8-10} 
			\textbf{M2OIE}  & 60.2 & 59.1 & \textbf{61.2} & 59.1 & 56.1 & \textbf{62.5} & 53.5 & 66.0 & 44.9     \\
			
			\textbf{milIE}        & \textbf{64.2}& 69.5 & 59.7 & \textbf{65.6} & 70.2 & 61.6 & 55.7 & 58.1 & 53.5                        \\
			\textbf{- DEP}     & 48.1 & 64.4 & 38.4     & 46.9&58.8 &39.0 & 45.0 & 62.0 & 35.3         \\ 
			\textbf{- NS}          & 59.1 &\textbf{ 75.7} & 48.5& 62.4 & \textbf{74.0} &54.0 &\textbf{59.5} & \textbf{66.2} & \textbf{53.9}              \\
			
			\bottomrule
		\end{tabular}
		\caption{\textsc{MiLLIE} performance comparison on \textsc{CaRB} lexical match (LM) benchmark. - DEP represents \textsc{milIE} trained and evaluated without dependency tags, -NS represents absence of negative sampling. \textsc{milIE} always outperforms M2OIE, except for the recall on the erroneous automatic translation of Spanish and Portuguese.}
		\label{tab:multilingual-carb}
	}
\end{table*}

\subsection{Setup}
\textbf{Baselines \& Training.} We compare \textsc{milIE} with both unsupervised and supervised baselines. Specifically we compare \textsc{milIE} with ClausIE, MinIE, Stanford-OIE, RNN-OIE, OIE6 \cite{Del2013clausie, gashteovski2017minie, Stanovsky2018supervised, angeli2015leveraging, Kolluru2020openie6} and Multi2OIE \cite{Ro2020multi2oie} on English. Multi2OIE is the only neural system capable of extracting triples from multiple languages and therefore it is the only available baseline for the non-English evaluations. 

We use the English RE-OIE2016 \cite{Zhan2020span} training dataset used in \cite{Ro2020multi2oie}. This training dataset contains n-ary extractions allowing \textsc{milIE} to be evaluated on both n-ary as well as binary extraction benchmarks. Evaluation on languages other than English is always zero-shot, i.e.,  the model is trained using only the English Re-OIE2016 dataset and tested on test set of the other languages.

\textbf{CaRB benchmark.} We use the \textsc{CaRB} benchmark introduced in \cite{Bhardwaj2019carb} for evaluating English OpenIE n-ary extraction. However, the \textsc{CaRB} benchmark also suffers from serious shortcomings due to its evaluation method based on token overlaps. For example, \cite{Gashteovski2021benchie} discovered that a simple OpenIE system that breaks the sentence into a triple at the verb boundary achieves $0.70$ recall and $0.19$ precision. This is problematic since it indicates that simply adding extraneous words to the extraction results in improved recall.

\textbf{BenchIE benchmark.} Due to the issues identified for CaRB, we also evaluate using BenchIE, which is an exhaustive fact based multilingual OpenIE benchmark proposed by~\cite{Gashteovski2021benchie}. BenchIE evaluates explicit binary extractions in English, Chinese, and German. BenchIE is accompanied by an annotation tool, AnnIE~\cite{Friedrich2021annie}, for extending the benchmark to additional languages. For Arabic, we translated 100 sentences from BenchIE-English to Arabic with the help of a native Arabic speaker and then extracted triples using AnnIE. Similarly for Galician we translated all 300 sentences to Galician with the help of a native Galician speaker who also annotated the dataset using AnnIE.

\textbf{Multilingual CaRB.} Additionally we also evaluate \textsc{milIE} on the Spanish and Portuguese multilingual CaRB datasets introduced in \citet{Ro2020multi2oie}. The lexical match evaluation used in this dataset has numerous shortcomings \cite{Bhardwaj2019carb}, however we include it for a fair comparison to \citet{Ro2020multi2oie}'s Multi2OIE system. The \textsc{CaRB} test set was translated to Spanish and Portuguese using the Google Translate API.
To investigate the quality of these automatic translations, we randomly sampled 100 sentences from the test sets and had them evaluated by native Spanish and Portuguese speakers. To our surprise we discovered that around 70 percent of the sentence or extraction translations were inaccurate. Table \ref{tab:errors} shows a few examples of the incorrect translations.
For an accurate and clean comparison with Multi2OIE we also cleaned up part of the Spanish test set by re-translating 149 sentences and their extractions in Spanish. These translations were done by native Spanish speakers.

On the \textsc{CaRB} English benchmark we use results for baselines reported in  \cite{Ro2020multi2oie} and \cite{Kolluru2020openie6}. For evaluating on BenchIE, we run all the baselines on the BenchIE English evaluation benchmark. For multilingual BenchIE we train Multi2OIE using the code and hyperparameters supplied in the paper. For hyperparameter tuning we use the \textsc{CaRB} English validation set and use the F1 scores obtained using the \textsc{CaRB} evaluation procedure for comparing models with different hyperparameters. The \textsc{milIE} model is trained using negative sampling and includes the dependency tag information and binarization. We use the spaCy dependency parser for obtaining dependency tags. We were unable to find a dependency parsing tool with universal dependencies for Arabic and therefore we did not use dependency tags for Arabic.
For BenchIE, \textsc{milIE} uses the binarization function described in Section \ref{sec:binarization}, but not for \textsc{CaRB} and lexical match because they evaluate n-ary extractions.

\subsection{Results}
\subsubsection{English}
In Table \ref{tab:eng}, we compare \textsc{milIE} with several unsupervised and supervised baselines in English on \textsc{CaRB} and BenchIE. \textsc{milIE} performs much better compared to other neural baselines on BenchIE. This is not the case for the \textsc{CaRB} dataset since \textsc{CaRB} penalizes compact extractions and rewards longer  extractions \cite{Gashteovski2021benchie}.
Although rule based systems like ClausIE and MinIE outperform neural systems, they cannot be used for languages other than English.

\subsubsection{Multilingual}

In Table \ref{tab:multilingual}, we compare \textsc{milIE} with Multi2OIE (M2OIE) on the multilingual BenchIE benchmark. \textsc{milIE} performs significantly better compared to Multi2OIE for all the languages. For German and Arabic both Multi2OIE and \textsc{milIE} perform significantly worse compared to the other languages. The presence of separable prefixes in German verbs which cannot be extracted using BIO tags results in low performance. The BIO tagging scheme assumes continuity of phrases which is absent for most German verbs present in predicates, resulting in extremely low recall. For Arabic, the low scores are due to the Verb-Subject-Object nature of the Arabic language along with the fact that subjects or objects can be expressed as part of the verb. This calls for additional research on framing OpenIE tasks for languages such as German and Arabic. \textsc{milIE} significantly outperforms Multi2OIE for Galician language which is closely related to Portuguese. Ablation results in Table \ref{tab:multilingual} also indicate the usefulness of adding the dependency tags, negative sampling, and the binarization mechanism.

In Table \ref{tab:multilingual-carb}, we compare \textsc{MiLIE} with Multi2OIE on the \textsc{CaRB} lexical match benchmark. \textsc{MiLIE}, without negative sampling works best for Spanish clean data. This is not due to the language, but due to the lexical match evaluation which rewards overly long extractions even if incorrect. Not using negative sampling sometimes improves recall which may improve F1 score. This is observed for the German benchmark. 


\begin{table}[]
	\ra{1.1}
	\footnotesize{
		\begin{tabular}{@{}lllllll@{}}
			\toprule
			\textbf{English}                                                      & \multicolumn{3}{c}{\textbf{CaRB-nary}}                  & \multicolumn{3}{c}{\textbf{BenchIE-binary}}               \\ \midrule
			\textbf{}                                                             & \textbf{F1} & \textbf{Prec.} & \textbf{Rec.} & \textbf{F1} & \textbf{Prec.} & \textbf{Rec.} \\
			\textbf{ClausIE}                                                      & 44.9        &        \textemdash            &    \textemdash             & 33.9       & 50.3              & 25.6           \\
			\textbf{MinIE}                                                        & 41.9        &  \textemdash                  &  \textemdash               & 33.7       & 42.9              & 27.8           \\ 
			\textbf{Stanford}                                                 & 23.0        &        \textemdash            &    \textemdash             & 13.0       & 11.1              & 15.7           \\
			\midrule
			\textbf{R-OIE}                                                      & 46.7        & 55.6               & 40.2            & 13.0       & 37.3              & 7.8            \\
			\textbf{S-OIE}                                                     & 49.4        & 60.9               & 41.6            &    \textemdash          &          \textemdash          &        \textemdash          \\
			\textbf{OIE6}                                                      & \textbf{52.7}        &       \textemdash             &  \textemdash               & 25.4       & 31.1              & 21.4           \\
			\textbf{M2OIE}                                                    & 52.3        & \textbf{60.9 }              & \textbf{45.8}            & 22.8       & 39.2              & 16.1           \\
			\textbf{milIE}                                                 &    45.0         &      48.6              &       41.8          & \textbf{27.9}       & \textbf{36.6}              & \textbf{22.4}           \\
			\textbf{-DEP}&     41.2        &      44.1              &         38.6        &       26.7      &        31.1            &      23.4           \\
			\textbf{-NS}   &     44.7        &    47.6                &      42.2           &       25.8      &       29.6             &  22.9\\      
			\textbf{-Bin}  &     \textemdash         &        \textemdash             &        \textemdash          &     27.7        &       34.6             &  23.1\\   
			\bottomrule
		\end{tabular}
	}
	
	\caption{\textsc{milIE} performance comparison on \textsc{CaRB} and BenchIE English benchmarks. \textsc{milIE} performs best out of all models on BenchIE. It performs worse compared to some model on CaRB, which is due to the CaRB evaluation scheme where overly long extractions are rewarded.}
	\label{tab:eng}
\end{table}

\begin{table}[]
	\centering
	\ra{1.1}
	\small{
		\begin{tabular}{@{}lllll@{}}
			\toprule
			\textbf{English} & \textbf{F1} & \textbf{Prec.} & \textbf{Rec.} & {$\mathbf{\Delta \ F1}$} \\ \midrule
			\textbf{\textsc{milIE}}        & 27.88               & 36.65     & 22.37 & \textemdash          \\
			\midrule
			\textbf{\textsc{milIE} + CO}           &  \textbf{29.71}           &  32.35       & \textbf{27.48}        & \textbf{+ 1.83 \%}       \\
			
			\bottomrule
		\end{tabular}
		\caption{Performance comparison of Hybrid \textsc{milIE} on English BenchIE. Here `+ CO' denotes system fused with extracted ClausIE Objects.}
		\label{tab:benchie_hybrid}
	}
\end{table}

\subsubsection{Hybrid OpenIE}
\textsc{milIE} can easily integrate any rule based system that extracts even a part of the triple. To evaluate this, we first simulate a system that only extracts the object and use \textsc{milIE} to extract other parts of the triple. We do this by employing ClausIE for extracting triples for the BenchIE English data and only use the object, discarding the rest of the triple.

The reason behind the choice of selecting object extraction from ClausIE is the fact that neural systems are not good at extracting objects \cite{Kolluru2020openie6}. This is also seen from additional experiments detailed in Section \ref{sec:analysis}. Table \ref{tab:benchie_hybrid} indeed confirms that combining rule based object extraction with \textsc{milIE} improves performance by over 6\% in F1 score. This showcases that \textsc{milIE}'s ability to integrate other systems can be a great advantage. 

\section{Analysis \label{sec:analysis}}

\begin{table}[]
	\centering
	\ra{1.1}
	\small{
		\begin{tabular}{@{}lllllll@{}}
			\toprule
			\textbf{F1-Score} & \textbf{EN} & \textbf{DE} & \textbf{ZH} & \textbf{AR} & \textbf{GL} & \textbf{ES} \\ \midrule
			\textbf{SPOA}     &     26.3         &  8.7           &  20.3           &  5.3           & 17.5 &    55.2                          \\
			\textbf{SOPA}     &     24.9         &  8.2           &   18.2          &    5.8         & 17.5 &     53.1                         \\
			\textbf{PSOA}     &     27.7         &   \textbf{8.8 }         &  19.5           &     5.0    &  17.5  &    51.4                           \\
			\textbf{POSA}     &     27.4         &     8.1        &   19.4          &   5.4          &  17.1  &   51.7                         \\
			\textbf{OSPA}     &     22.4         &     8.0        &   17.1          &   5.7          &  15.3  &         45.5                  \\
			\textbf{OPSA}     &      22.2        &     7.9        &   17.5          &   6.4          &  15.2   &   47.9                       \\
			\textbf{DYN}  &        26.9     &       \textbf{9.0}      &      19.5       &    4.9         &  17.5   & 51.0                         \\ 
			\textbf{WF}       &    \textbf{27.9}          &     8.5        &   \textbf{20.5}          &      \textbf{7.3}       & \textbf{18.3} & \textbf{55.7 }                          \\
			\bottomrule
		\end{tabular}
		\caption{Comparison between different decoding schemes. WF represents water filling and DYN the dynamic setting.}
		\label{tab:decoding}
	}
\end{table}

We would like to analyze that the ability of \textsc{milIE} to extract triples using different extraction patterns results in improved performance on multilingual data. For this, we compare \textsc{milIE} with the water filling aggregation against \textsc{milIE} with different extraction pathways.
We also compared with a dynamic decoding scheme where \textsc{milIE} chooses a decoding pathways based on the sentence. To do this we split a part of the English training set and for each sentence in the split we record the extraction pathway that provides the best F1 score \textsc{milIE} as per \textsc{CaRB} evaluation. We then use this as training data for training another mBERT model which classifies each sentence in one of the six classes where each class represents an extraction pathway. 

Table \ref{tab:decoding} details the performance for different extraction schemes. All the extraction schemes except WF, use only one pathway. DYN provides mixed results across the different languages - for German it is the best approach, whereas for Arabic it is the worst. In contrast, the combination of multiple pathways allows to performing much better than the other approaches on all languages, except German. 
This demonstrates that combining triple extraction from multiple pathways is better than any single pathway, which in turn confirms that extracting triples repeatedly from the same sentence using multiple extraction pathways is more profitable than using a single extraction pathway.

\begin{figure}
	\centering
	\includegraphics[scale=0.48]{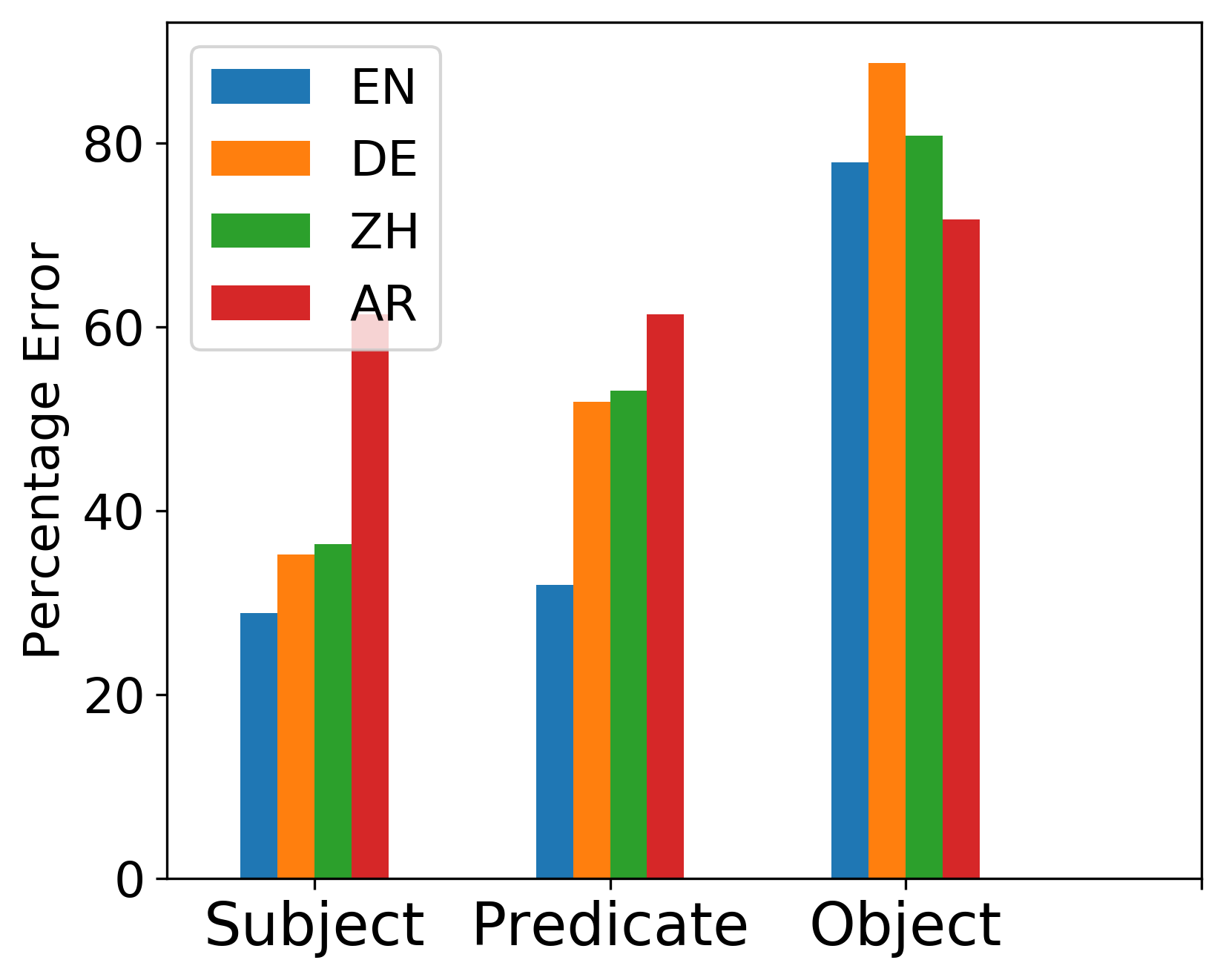}
	\caption{Percentage error contribution due to incorrect subject, predicate or object for EN, DE, ZH and AR. Most errors occur in the object. }
	\label{fig:errors}
\end{figure}

Additionally, Table \ref{tab:decoding} provides an interesting insight: predicate first seems to be the best, followed by subject first and then object first for languages other than Arabic. This also shows how the difficulty of extracting triple slots using transfer learning from English varies with the target language.

Table \ref{tab:decoding} suggests that predicates are easier to extract leading to lesser number of errors propagated in the prediction chain. We suspect that this could result from differences in linguistic variability. To test our hypothesis we measured the entropy of the distribution of dependency and part-of-speech tags in the predicate, subject and object slots in the BenchIE English and the multilingual test sets. Results shown in Table \ref{tab:ent} suggest that linguistic complexity of objects is higher than those of predicates and subjects. 


This is also confirmed in Figure \ref{fig:errors}, where we plot the extraction errors in either subject, predicate or objects among incorrectly extracted triples. Most errors result from extracting incorrect objects compared to predicates and subjects. The percentage sum does not add to hundred because an incorrect triple can contain errors in more than one slot.


\begin{table}[]
	\centering
	\ra{1.1}
	\small{
		\begin{tabular}{@{}lllllll@{}}
			\toprule
			& \multicolumn{2}{c}{\textbf{Subject}} & \multicolumn{2}{c}{\textbf{Predicate}} & \multicolumn{2}{c}{\textbf{Object}} \\  \cmidrule(lr){2-3} \cmidrule(lr){4-5} \cmidrule(lr){6-7}
			\textbf{}   & \textbf{DEP}      & \textbf{POS}     & \textbf{DEP}       & \textbf{POS}      & \textbf{DEP}     & \textbf{POS}     \\
			\cmidrule(lr){2-2} \cmidrule(lr){3-3} \cmidrule(lr){4-4} \cmidrule(lr){5-5} \cmidrule(lr){6-6} \cmidrule(lr){7-7}
			\textbf{EN} &       1.719            &   1.588                &     2.443               &    1.831               &      2.286            &         1.861         \\
			\textbf{ZH} &      2.464             &       1.827           &          2.497          &      1.476             &   2.602               & 1.943          \\
			\textbf{DE} &       1.587            &       1.567          &    1.811            &     1.457            &    2.115     & 2.095          \\ 
			\bottomrule
		\end{tabular}
		\caption{Entropy of dependency and part of speech tags for subject, predicate and objects in BenchIE test data. Objects exhibit the highest entropy which indicates their higher complexity.}
		\label{tab:ent}
	}
\end{table}
\section{Related Work}
OpenIE systems largely come in two flavors, (1) unsupervised OpenIE systems that use fine grained rules based on dependency parse trees \cite{Del2013clausie, gashteovski2017minie, lauscher2019minscie}, and (2) supervised neural OpenIE systems, trained end-to-end with large training datasets \cite{Stanovsky2018supervised, Ro2020multi2oie, Kolluru2020openie6}.  Neural OpenIE systems characterize OpenIE as either a sequence tagging task \cite{Stanovsky2016EMNLP, Ro2020multi2oie}, span prediction task or a sequence generation task \cite{Kolluru2020imojie}. However all these prior approaches extract a triple in a single step, which does not allow us to study the effect of extracting a specific slot and its effect on extracting the rest of the triple. 



Neural generative approaches to OpenIE use sequence-to-sequence models with a copy mechanism for generating triples \cite{Sun2018Logician, Kolluru2020imojie}. The copy mechanism needs to be learned and is often a source of errors. 
A series of alternative approaches cast OpenIE as a sequence tagging task where each token is tagged as subject, predicate or object using a BIO like tagging scheme \cite{Stanovsky2018supervised, Ro2020multi2oie, Kolluru2020openie6}.
In these systems, all triple slots are extracted simultaneously and it is therefore not possible to condition on easier slots.

More closely related to our work is SpanOIE \cite{Zhan2020span} and Multi2OIE \cite{Ro2020multi2oie}, which first extracts the predicate and then all additional arguments. Like us, Multi2OIE \cite{Ro2020multi2oie} addresses multilinguality by leveraging a pretrained BERT model \cite{Devlin2019bert} for transfer learning. In contrast, through our iterative nature, it is possible to enrich the extractions in other languages if rule based models or other models (e.g. NER recognizers) exist to provide input for a triple slot.
IMOJIE \cite{Kolluru2020imojie} iteratively extracts entire triples from a sentence: first a triple is extracted, which is added to the input to extract the next triple. In contrast, our work iteratively extracts the slots of a single triple, which allows us to condition on the easier slots and therefore obtain higher quality triples.  \cite{Kolluru2020openie6} propose OpenIE6, a BERT based system, with iterative grid labelling and linguistic constraint based training. Such lingusitic constraints with soft penalties cannot be readily ported to other languages since such constraints use head verb based heuristics. Consequently OIE 6 is evaluated only on English.

\section{Conclusion}
We introduced \textsc{milIE}, a modular \& iterative multilingual OpenIE system. We confirmed our hypothesis that it is beneficial to extract triple slots iteratively which allows us to extract easier slots first. 
Our experiments on English as well as five low resource languages uncovered that, with the exception of Arabic, triples are easier to extract if the predicate is extracted first followed by the subject and object. More importantly we discovered that extracting triples using multiple extraction pathways is superior than the standard single extractions especially in the multilingual setting. We also demonstrated how \textsc{milIE} can be combined seamlessly with rule based systems for improving performance. Although our experiments were focused on the OpenIE task, we believe that the insights gained can be translated to other information extraction tasks with coupled extractions. We plan to explore such connections in the future.

\section{Acknowledgements}
We would like to thank Karin Letícia Mariani for checking and enumerating the errors in the \textsc{CaRB} Portuguese language dataset.

\bibliographystyle{acl_natbib}
\bibliography{milIE}
\appendix
\section{Appendix \label{sec:appendix}}

\subsection{Training Details}

\textsc{milIE} is expected to predict slots iteratively conditioned on prior extracted slots of a triple, therefore it needs to be trained with similar examples.  Exhaustively listing all possible combinations of prior extracted slots and slots to be extracted is prohibitively expensive. Therefore we use a sampling procedure that ensures the model sees a variety of combinations during training.

For every example in the Re-2016 training dataset we do the following
\begin{enumerate}
	\item Sample an slot as target (for extraction) with the following probabilities (subject: 1/3, predicate: 5/12, object: 5/12)
	\item Sample two slots, one that is assumed to be extracted and other the target that needs to be extracted conditioned on the first. 
	\item Sample three slots, first two assumed to be extracted and the third is the target conditioned on first two.
	\item If the example contains n-ary arguments, the subject, predicate and object are assumed to be extracted and the n-ary arguments are treated as targets.
\end{enumerate}

When a slot is sampled for target extraction, all instances of the slot are expected to be extracted. For example, if the target is the subject and if the example consists of multiple subjects then the targets are multiple subjects. However the sampled slots assumed to be extracted must be single instances, and if there are multiple instances, then each instance is considered for conditioning one after the other. Table \ref{table:train} details the sampling probabilities for two and three slots. The sampling probabilities were not tuned, but rather chosen based on heuristics. Post sampling, we obtain training dataset with about 5 and a half million examples.
\\ \\
\textbf{Negative Sampling}

We provide \textsc{milIE} with negative samples during training for reducing error amplification arising out of iterative prediction. In this case the target is always blank, i.e., all the tokens are marked as 'outside'. Thus the sampling revolves around creating incorrectly extracted slots. We sample negatives for every example in the training data and then select $k$ negative samples uniformly at random. $k$ is treated as a hyperparameter.

Table \ref{table:ns} provides the sampling probabilities for different slot arrangements. We use three corruption procedure for generating incorrectly marked slots, namely, invert, randomize and switch. The invert method consist of swapping the extracted slot with the target slot. For example, if the extracted slot is subject and target slot is object, then the object is marked as subject. The randomize method consists of choosing a random span of tokens near the actual slot. Finally the switch method involves switching one of the extracted slot with a slot from another triple associated with the sentence. For example, in the case of (subject, object), the object of this triple is switched with an object of another triple associated with the same sentence. It is possible that the same subject maybe associated with the new object as well. We check if this is true, and if true we filter out such positives.

\begin{table}[]
	\begin{tabular}{@{}llc@{}}
		\toprule
		\textbf{Extracted Slots} & \textbf{Target Slot} & \textbf{Probability} \\ 
		\cmidrule(r){1-1} \cmidrule(r){2-2} \cmidrule(r){3-3}
		subject                      &       object               &          3/12            \\
		subject                     &       predicate               &           1/12           \\
		object                     &    subject                  &              2/12        \\
		object                     &    predicate                  &            1/12          \\
		predicate                     & subject                     &           2/12           \\
		predicate                     &     object                 &            3/12          \\
		\midrule
		(subject, object)                     &   predicate                   &  2/12                    \\
		(subject, predicate)                     &    object                  &     6/12                 \\
		(object, predicate)                     &     subject                 &   4/12                   \\
		\bottomrule
	\end{tabular}
	\caption{Sampling Probabilities for training data.}
	\label{table:train}
\end{table}

\begin{table}[]
	\small{
		\begin{tabular}{@{}llll@{}}
			\toprule
			\textbf{Extracted Slots} & \textbf{Target} &\textbf{Corruption} & \textbf{Prob.} \\ 
			\cmidrule(r){1-1} \cmidrule(r){2-2} \cmidrule(r){3-3} \cmidrule(r){4-4}
			subject         &  object & Invert         &    1/12                  \\
			object     & predicate & Invert             &     3/12                 \\
			predicate             & subject & Randomize           &     2/12                 \\
			(subject, object)       & predicate & Switch       &       1/12               \\
			(subject, predicate)     & object & Switch                    & 3/12                     \\
			(predicate, object)     &  subject & Switch     &       2/12               \\
			\bottomrule
		\end{tabular}
		\caption{Negative Sampling Probabilities.}
		\label{table:ns}
	}
	
\end{table}

\subsection{Hyperparameter Tuning}
We train and evaluate \textsc{milIE} on an NVIDIA Titan RTX with 24 GB GPU RAM. The training is done for a maximum of two epochs and each epoch takes about 9-10 hours. The maximum sentence length using the English train and validation dataset is found to be about 100. Due to the addition of extracted triple element markers we allow a slack of 20 tokens, thus fixing the maximum sentence length to 120. We use a maximum possible batch size that fits inside the GPU, which results in batch size of 192. We use \textsc{Adam} \cite{adam} as the optimizer with linear warmup and tune the learning rate. The linear warmup fraction is fixed at $0.1$. We also treat the number of negative samples, $k$, as a hyperparameter and tune it. We choose the best hyperparameters based on the F1 score. Table \ref{table:tune} provides details on the recall scores for every hyperparameter arrangement.

\begin{table}[]
	\begin{tabular}{@{}lll@{}}
		\toprule
		\textbf{Num. NS (k)} & \textbf{Learning Rate} & \textbf{F1} \\ \midrule
		0             & $1 \times 10^{-5}$   &       39.88          \\
		0             & $3 \times 10^{-5}$   &        44.70         \\
		0             & $9 \times 10^{-5}$   &       40.76          \\ \midrule
		10K           & $1 \times 10^{-5}$   &       43.45         \\
		10K           & $3 \times 10^{-5}$   &        47.03         \\
		10K           & $9 \times 10^{-5}$   &        47.19         \\ \midrule
		100K          & $1 \times 10^{-5}$   &         48.03       \\
		100K          & $3 \times 10^{-5}$   &         47.30        \\ 
		100K          & $9 \times 10^{-5}$   &       45.87          \\ \midrule
		1M            & $1 \times 10^{-5}$   &            46.01     \\
		1M            & $3 \times 10^{-5}$   &       46.16          \\
		1M            & $9 \times 10^{-5}$   &            45.26    \\
		\bottomrule
	\end{tabular}
	\caption{Hyperparameter Tuning}
	\label{table:tune}
\end{table}

\begin{figure*}
	\centering
	\includegraphics[scale=0.16]{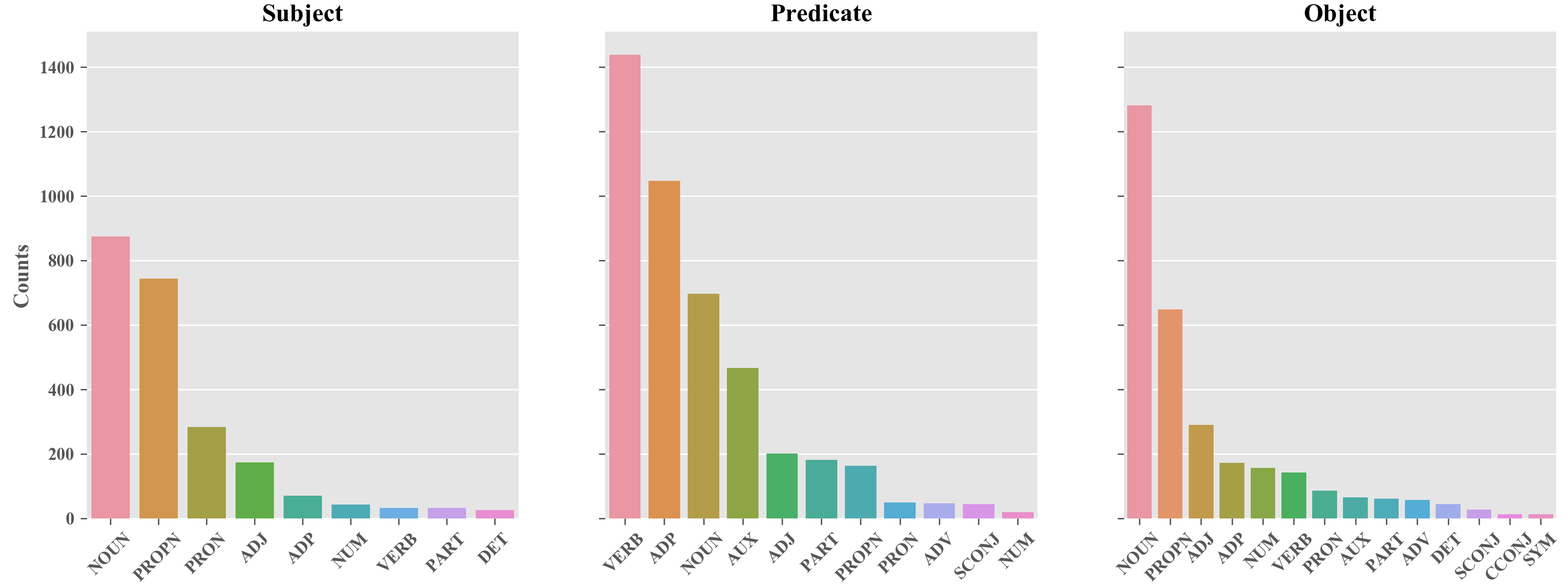}
	\caption{Distribution of the Part of Speech tags in subject, predicates and object tokens of triples in BenchIE English test data. }
	\label{fig:pos}
\end{figure*}

\end{document}